\documentclass[fleqn,10pt]{wlscirep}
\usepackage[utf8]{inputenc}
\usepackage[T1]{fontenc}
\usepackage{multirow}

\title{Predicting Goal-directed Attention Control Using Inverse-Reinforcement Learning}

\author[1,2,*]{Gregory J. Zelinsky}
\author[1]{Yupei Chen}
\author[1]{Seoyoung Ahn}
\author[1]{Hossein Adeli}
\author[2]{Zhibo Yang}
\author[2]{Lihan Huang}
\author[2]{Dimitrios Samaras}
\author[2]{Minh Hoai}
\affil[1]{Department of Psychology, Stony Brook University, Stony Brook, NY 11794 USA}
\affil[2]{Department of Computer Science, Stony Brook University, Stony Brook, NY 11794 USA}

\affil[*]{gregory.zelinsky@stonybrook.edu}

\begin{abstract}
Understanding how goal states control behavior is a question ripe for interrogation by new methods from machine learning. These methods require large and labeled datasets to train models. To annotate a large-scale image dataset with observed search fixations, we collected 16,184 fixations from people searching for either microwaves or clocks in a dataset of 4,366 images (MS-COCO). We then used this behaviorally-annotated dataset and the machine learning method of Inverse-Reinforcement Learning (IRL) to learn target-specific reward functions and policies for these two target goals. Finally, we used these learned policies to predict the fixations of 60 new behavioral searchers (clock = 30, microwave = 30) in a disjoint test dataset of kitchen scenes depicting both a microwave and a clock (thus controlling for differences in low-level image contrast).  We found that the IRL model predicted behavioral search efficiency and fixation-density maps using multiple metrics. Moreover, reward maps from the IRL model revealed target-specific patterns that suggest, not just attention guidance by target features, but also guidance by scene context (e.g., fixations along walls in the search of clocks). Using machine learning and the psychologically-meaningful principle of reward, it is possible to learn the visual features used in goal-directed attention control.
\end{abstract}

\begin{document}

\flushbottom
\maketitle
%
%
\thispagestyle{empty}


\section*{Introduction}


Ever since Yarbus’ seminal demonstration of how a goal can control attention~\cite{yarbus1967eye}, understanding goal-directed attention control has been a core aim of psychological science. This focus is justified. Goal-directed attention underlies everything that we \textit{try} to do, making it key to understanding cognitively-meaningful behavior. Like Yarbus, we too demonstrate goal-directed control of eye-movement behavior, but here these overt attention movements are made by a deep-network model that has learned different goals. 

Three factors distinguish our approach from previous work. First, it is image-computable and uses learned, rather than handcrafted, features. Our model therefore inputs an image but is not told anything about its features ("vertical", "clock", etc.), which all must be learned. This factor distinguishes the current model from most others in the behavioral literature on attention control~\cite{wolfe1994guided, zelinsky2008theory, bundesen1990theory}, and makes our approach more aligned with recent computational work.~\cite{zhang2018finding, zelinsky2019benchmarking} Second, the goal-directed behavior that we study is categorical search, the visual search for any exemplar of a target-object category.~\cite{schmidt2009search,eimer2014neural,zelinsky2013modeling} We adopt this paradigm because categorical search is the simplest (and therefore, best) goal-directed behavior to computationally model---there is a target-object goal and the task is to find it. A third and unique contribution of our approach is that we predict categorical-search fixations using a policy that was learned, through many observations of search-fixation behavior during training, to maximize the goal-specific receipt of reward. Using inverse-reinforcement learning (IRL), we obtain these reward functions and use them to prioritize spatial locations to predict the fixations made by new people searching for the learned target categories in new images. Doing this required the creation of a search-fixation-annotated image dataset sufficiently large to train deep-network models (see Methods). We show that this model successfully captured several patterns observed in goal-directed search behavior, not the least being the guidance of overt attention to the target-category goals.

\subsection*{Inverse-Reinforcement Learning}
IRL is an imitation-learning method from the machine-learning literature that learns, through observations of an expert, a reward function and policy for mimicking expert performance. We extend this framework to goal-directed behavior by assuming that the image locations fixated by searchers constitute the expert performance that the model learns to mimic. The specific IRL algorithm that we use is Generative Adversarial Imitation Learning (GAIL~\cite{ho2016generative}), which makes reward proportional to the model’s ability to generate State-Action pairings that imitate observed State-Action pairings. Here, the Action is a shift of fixation location in a search image (the model’s saccade), and the State is the search context (all the information available for use in the search task). The State includes, but is not limited to, the visual features extracted from an image and the learned visual representation of the target category. Over training, and through the greedy maximization of total-expected reward, the model learns a Policy for mapping States to Actions that can be used to predict new Actions (saccades) given new States (search images).

\section*{Methods}
\label{sec:modelmethods} 

\subsection*{Model Methods}
\begin{figure}[!hbt]
\centering
\includegraphics[width=1\textwidth,height=0.95\textheight,keepaspectratio]{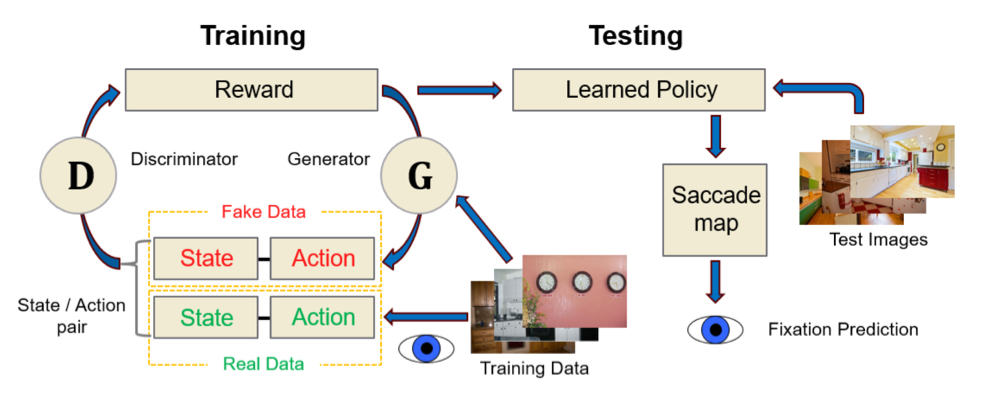}
\caption{The model's adversarial imitation learning algorithm. During training it learned from fixation-annotated images a reward function and policy for predicting new search fixations in unseen test images.}
\label{fig:pipeline}
\end{figure}

The IRL model framework is illustrated in Fig.~\ref{fig:pipeline}. Model training can be conceptualized as a Policy Generator (G) and a Discriminator (D) locked in an adversarial process~\cite{ho2016generative}. The Generator generates fake eye movements (Actions) with the goal of fooling the Discriminator into believing that these actions were made by a person, while the Discriminator's goal is to discriminate the real eye movements from the fake. More specifically, the Generator consists of an Actor-Critic model~\cite{konda2000actor}
that learns a policy for maximizing total expected reward over all possible sequences of fixations, with greater reward given to the Generator when it produces person-like actions that the Discriminator miss-classifies as real (the logarithm of the Discriminator output). This reward-driven adversarial process plays out during training using Proximal Policy Optimization (PPO)~\cite{schulman2017proximal}, with the result being a Generator that becomes highly adept at imitating the behavioral fixations made during categorical search. At testing, this learned Policy for mimicking people's categorical search fixations is used to predict the fixation behavior of new people searching for the same target categories in new images. These fixation predictions are quantified by what we call a \textit{saccade map}, which is a priority map reflecting the total reward expected if saccades were to land at all the different locations in an image input.

\subsection*{States and Actions: Cumulative Movements of a Foveated Retina}
\begin{figure}[!hbt]
\centering
\includegraphics[width=1\textwidth,height=0.95\textheight,keepaspectratio]{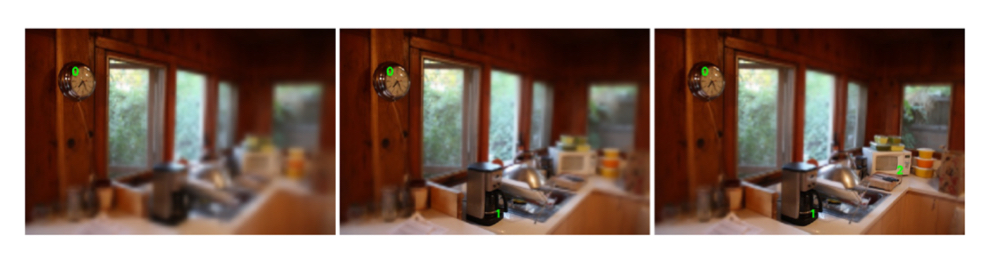}
\caption{The formation of a cumulative retina-transformed image over the first three fixations (0,1,2).}
\label{fig:retimages}
\end{figure}

Broadly speaking, the State is the internal visual representation that is used for search, and a big part of this are the features extracted from the image input. To obtain a robust core State representation we pass each image through a pre-trained ResNet-50~\cite{He-et-al-ICCV15} to get a reasonably-sized feature map output (1024x10x16). However, human search behavior is characterized by movements of a foveated retina, and each of these search fixations dramatically changes the State by re-positioning the high-resolution fovea in the visual input. We captured this fixation-dependent change in State in two steps. First, we gave the IRL model a simplified foveated retina. We did this using the method from Geisler and Perry~\cite{perry2002gaze} to compute a retina-transformed version of the image input (\textit{ReT-image}), which in our implementation is an image having high resolution within a central $3^\circ$ "fovea" (32x32 in the resized 512x320 pixel image) but is blurred outside of this fovea to approximate the loss of resolution that occurs with increasingly eccentric viewing in the visual periphery. Second, we accumulate these high-resolution foveal views, each a different ReT-image, over 6 new fixations in a process that we refer to as \textit{cumulative foveation}. With each new "eye movement", the fovea is re-positioned in the image, thereby progressively de-blurring what was an initially fairly blurred visual input. Note that by adopting this cumulative-foveation State encoder we are not suggesting that people have a similar capacity to maintain high-resolution visual information once the fovea moves on, and indeed this is known not to be the case~\cite{irwin1996integrating}. Rather, we used this fixation-by-fixation State encoder simply as a tool to integrate a dynamically changing State into the IRL method. Figure~\ref{fig:retimages} shows cumulative ReT-images obtained at three successive fixation locations (0,1,2) for a sample scene, with the 7-fixation sequence of these images comprising a dynamic State representation that is input to the IRL model. The pre-trained ResNet-50 was dilated and fine-tuned on ReT-images prior to this State encoding.

The IRL model learns to associate States with Actions, but these Actions must also be defined in some space. We obtain an Action space by first resizing a ReT-image input to 512x320 pixels, which we then discretize into a 10x16 grid of 32x32 pixel cells. The center of each cell becomes a potential fixation location, a computational necessity imposing a resolution limit on the model's oculomotor behavior. For each of the 6 new fixations generated by the model, the cumulative ReT-image input is prioritized by the saccade map and one of the 160 possible grid locations is selected for an eye movement.

\subsection*{The Microwave-Clock Search Dataset}
\label{sec:dataset}
The currently most predictive models of complex fixation behavior are in the context of a free-viewing task, where the best of these models (e.g., DeepGaze II~\cite{kummerer2017understanding}) are trained on SALICON~\cite{jiang2015salicon}. SALICON is a crowd-sourced dataset consisting of images that were annotated with human mouse clicks indicating salient image locations. Without SALICON, DeepGaze II and models like it would not have been possible, and our understanding of free-viewing behavior, widely believed to reflect bottom-up attention control (i.e., control solely by features extracted from the visual input), would be diminished. To date, however, there has been no comparable dataset for categorical search, and this has hindered the computational modeling of goal-directed attention control. Those suitably-sized and fixation-annotated image datasets that do exist either did not use a standard search task ~\cite{mathe2014actions,papadopoulos2014training}, 
 used a search task but had people search for multiple targets simultaneously~\cite{gilani2015pet}, or used only one target category (people)~\cite{ehinger2009modelling}. Here we introduce the Microwave-Clock Search (MCS) dataset, which is now among the largest datasets of images that have been annotated with goal-directed fixations. The MCS dataset makes it possible to train deep-network models on human search fixations to predict how people will move their attention in the pursuit of different target-object goals. 

\begin{figure}[!hbt]
\centering
\includegraphics[width=1\textwidth,height=0.95\textheight,keepaspectratio]{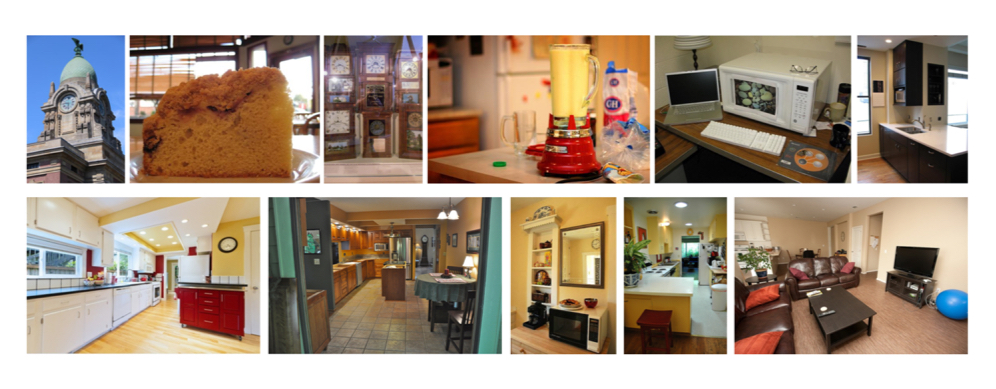}
\caption{Representative training images (top) and testing images (bottom) in the MCS dataset.}
\label{fig:cocoimages}
\end{figure}

Half of the MCS dataset consists of COCO2014 images~\cite{lin2014microsoft} depicting either a microwave or a clock (based on COCO labels), from which we created disjoint training and testing datasets. In selecting the training images we excluded scenes depicting people and animals (to avoid attention biases to these categories), and digital clocks in the case of the clock target category. This latter constraint was introduced because the features of analog and digital clocks are very different, and we were concerned that this would introduce unwanted variability in the search behavior. No additional exclusion criteria were used to select the training images, with our goal being to include as many images for training as possible. These criteria left 1,494 analog clock images and 689 microwave images, which we should note varied greatly in terms of their search difficulty (see Fig.~\ref{fig:cocoimages}, top). Selection of the test images was more tightly controlled, resulting in the test dataset being far smaller (n=40). In addition to the exclusion criteria used for the training images, test images were further constrained to have: (1) depictions of \textit{both a microwave and a clock} (enabling different targets to be designated in the identical images, the perfect control for differences in bottom-up saliency), (2) only a single instance of the target, (3) a target area less than 10\% of the image area, and (4) targets that do not appear at the image’s center (no overlap between the target and the center cell of a 5x5 grid). The latter two criteria were aimed at excluding really large targets or targets appearing too close to the center starting-gaze position, with the goal of both being to achieve a moderate level of search difficulty (see Fig.~\ref{fig:cocoimages}, bottom).

\begin{figure}[!hbt]
\centering
\includegraphics[width=1\textwidth,height=0.95\textheight,keepaspectratio]{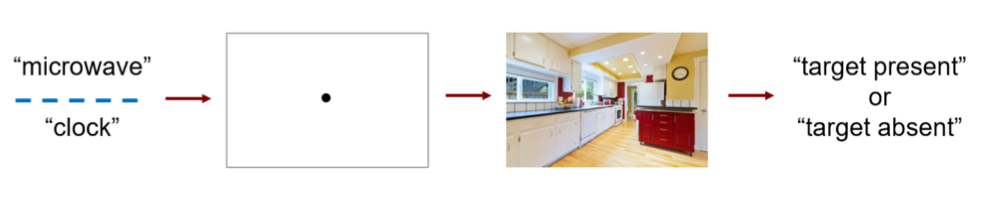}
\caption{The categorical search paradigm used for behavioral data collection.}
\label{fig:behaviorexp}
\end{figure}

The above-described selection criteria were specific to target-present (TP) images, but an equal number of target-absent (TA) images (n=2183) were selected as well so as to create a standard TP versus TA search context. These images were selected randomly from COCO, with the constraints that: (1) none depicted the target, and (2) all depicted at least two instances of the target category's siblings. COCO defines the siblings of a microwave to be: ovens, toasters, refrigerators, and sinks, all under the parent category of “appliances”. Clock siblings are defined as: books, vases, scissors, hairdryers, toothbrushes, and teddy bears, under the parent category of “indoor”. Sibling membership was used as a selection criterion so as to discourage TA responses from being based on scene type (e.g., a street scene is unlikely to contain a microwave), and this criterion seemed to work well; the overwhelming majority of selected TA scenes were kitchens that did not depict a target. 

The large size of the training dataset (4366 images) required data collection to be distributed over groups of searchers. Each microwave training image was searched by 2-3 people (n=27); each clock training image was searched by 1-2 people (n=26). After removing incorrect trials and TP trials in which the target was not fixated (it is not desirable to train on these), 16,184 search fixations remained for model training. Test images were each searched by a new group of 60 participants, 30 searching for a microwave target and the other 30 searching the same images for a clock target in a between-subjects design. To achieve a power and effect size of .8, based on a t-test comparing target guidance to chance (see Fig.~\ref{fig:cumprob}), we determined that a sample of 25 participants per target condition would be adequate. However, we chose to test 30 participants per condition in case of loss due to attrition or unusable eye-tracking data.

\subsection*{Behavioral Search Procedure}
A standard categorical search paradigm was used for both training and testing (see Fig.~\ref{fig:behaviorexp}). TP and TA trials were randomly interleaved within target type, and searchers made a speeded TP or TA manual response terminating each trial. Search display visual angles were $54^\circ \times 35^\circ$ for testing; for training angles ranged from $12^\circ \times 28.3^\circ$ in width and $8^\circ \times 28.3^\circ$ in height. Eye position was sampled at 1000 Hz using an EyeLink 1000 (SR Research) in tower-mount configuration (spatial resolution $0.01^\circ$ rms). All participants provided informed consent in accordance with policies set by the institutional review board at Stony Brook University responsible for overseeing research conducted on human subjects. 

\section*{Results}
\label{results}

\subsection*{Search Behavior}
Table~\ref{tab:dataset} provides the mean button-press errors and the average number of fixations made before the button-press response (which includes the starting fixation) on correct search trials. Note that the roughly doubled error rates in the training data should be interpreted with caution, as many of these errors were due to incorrectly labelled target-object regions in COCO that would cause errors given correct search judgments. Rather than correcting these mislabelled objects (which would be changing COCO), we instead decided to tolerate an inflated error rate and to exclude these error trials from all analyses and interpretation. 

\begin{table}[!hbt]
\centering
\begin{tabular}{|l|l|l|l|l|l|}
\hline
\multirow{2}{*}{} & \multirow{2}{*}{Target Category} & \multicolumn{2}{l|}{Training Dataset} & \multicolumn{2}{l|}{Testing Dataset} \\ \cline{3-6} 
 &  & Error (\%) & Mean (SD) Fixations & Error (\%) & Mean (SD) Fixations \\ \hline
\multirow{2}{*}{target-present} & microwave & 18 & 5.46 ($\pm2.6$) & 9 & 6.76 ($\pm2.1$) \\ \cline{2-6} 
 & clock & 15 & 4.52 ($\pm3.5$) & 6 & 5.33 ($\pm1.8$) \\ \hline
\multirow{2}{*}{target-absent} & microwave & 8 & 7.95 ($\pm4.1$) & 4 & 14.36 ($\pm2.5$) \\ \cline{2-6} 
 & clock & 10 & 11.14 ($\pm6.8$) & 5 & 15.85 ($\pm2.3$) \\ \hline
\end{tabular}
\caption{Summary statistics showing mean errors and number of search fixations in the Microwave-Clock Search dataset.}
\label{tab:dataset}
\end{table}

\begin{figure}[!hbt]
\centering
\includegraphics[width=0.7\textwidth,height=0.95\textheight,keepaspectratio]{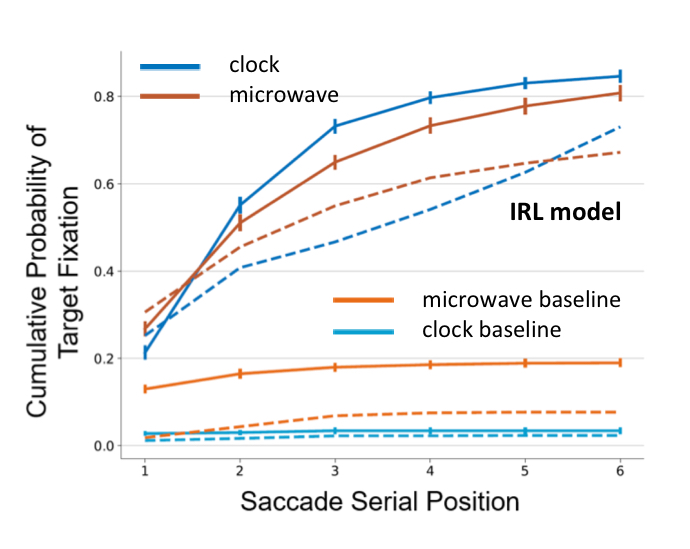}
\caption[short]{Cumulative probability of fixating the microwave (red) or clock (blue) target on target-present trials for behavioral participants (solid lines) and the model (dashed lines). Bottom lines are object-based random baselines (see text for details).}
\label{fig:cumprob}
\end{figure}

Focusing first on the TP test data, Figure~\ref{fig:cumprob} plots the cumulative probability of fixating the target with each saccade made during search. The central behavioral data pattern (solid lines) is that attention, as measured by overt gaze fixation, is strongly guided to both the microwave and clock targets. This guidance is evidenced by the fact that 24\% of the initial saccades landed on targets (averaged over microwaves and clocks). This probability of target fixation is well above chance, which we quantified using two object-based chance baselines consisting of: (1) the probability of fixating the clock when searching for a microwave (clock baseline), and (2) the probability of fixating the microwave when searching for a clock (microwave baseline). We confirmed above-chance target guidance by comparing the slopes of regression lines fit to the target and baseline data (microwave: target slope = 0.15, baseline slope = 0.03, t(58) = 26.31, p = 6.20e-34 < .001; clock: target slope = 0.17, baseline slope = 0.004, t(58) = 52.65, p = 1.14e-50 < .001). 
Also evident from this analysis is the importance of the first six saccades made during the search tasks. If the target was going to be fixated, it is highly likely that this would happen by the sixth eye movement. Collectively, these results indicate that there are strong microwave and clock guidance signals in the behavioral test data to predict.

\subsection*{IRL Model}
\begin{figure}[!hbt]
\centering
\includegraphics[width=1\textwidth,height=0.95\textheight,keepaspectratio]{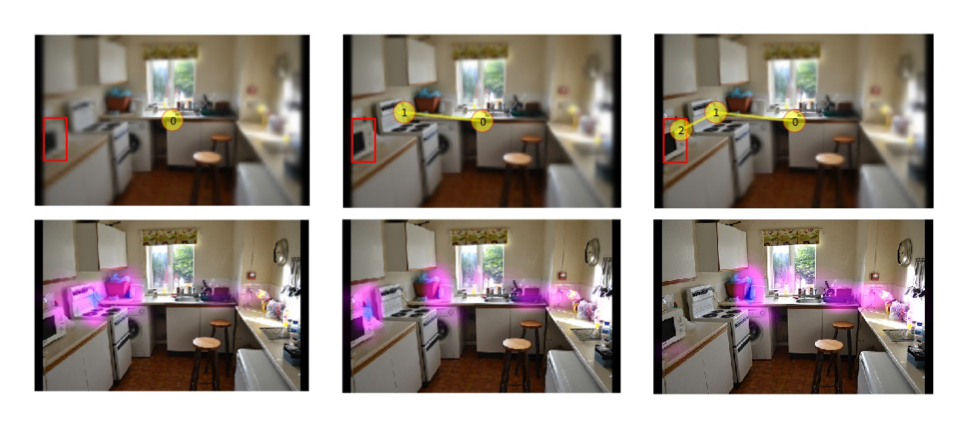}
\caption[short]{Cumulative ReT-images (top row) and corresponding saccade maps (bottom row) for the initial and first two new fixations (left to right) made by the model in a microwave search task.}
\label{fig:saccademap}
\end{figure}

To determine whether the model’s behavior is reasonable, we conducted two initial qualitative analyses. The top row in Figure~\ref{fig:saccademap} shows cumulative ReT-images for the starting fixation (0 in the yellow scanpath) and the fixations following the first two saccades (1, 2). Note that the left ReT-image, because it was computed based on a center initial fixation position, is blurred on both the left and right sides. The middle and right ReT-images were computed based on the landing positions of the first and second saccades, respectively. The microwave target is indicated in each panel by the red box. The bottom row shows the saccade maps corresponding to these ReT-images, where a bluer color indicates greater total reward expected by moving fixation to different image locations. The model initially expected the greatest total reward by fixating the stove (left saccade map), but after that saccade, and the resulting change in State (top middle), the model then selected the microwave target as the location offering the greatest expected reward (bottom middle), which was fixated next (right panels). Note that the model, because it was forced to make six saccades (discussed below), continued to prioritize space even after fixating the target. This qualitative analysis shows that the model learned an association between a State (which includes the features of a microwave) and an Action, and this enabled it to guide its fixations during the search of a new image for this target-category goal.

\begin{figure}[!hbt]
\centering
\includegraphics[width=1\textwidth,height=0.95\textheight,keepaspectratio]{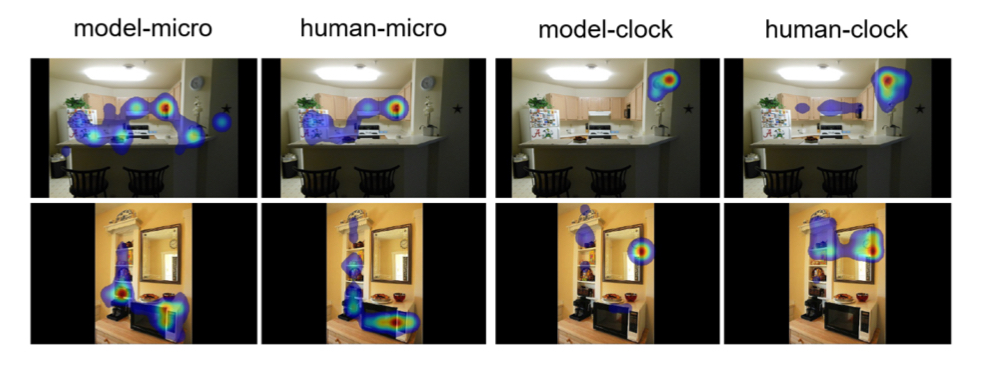}
\caption[short]{Model and behavioral Fixation Density Maps computed for microwave (left four) and clock (right four) searches in two trials (top, bottom).}
\label{fig:fdm}
\end{figure}

Figure~\ref{fig:fdm} shows another qualitative evaluation, this time comparing Fixation-Density Maps (FDMs) from people searching for a microwave (n=30) or a clock (n=30) to FDMs generated by the model (sampling from probabilistic policy) as it searched for the same targets in the same two test images. In both examples, the model and behavioral searchers efficiently found the target (bright red). More interesting, however, is that they both searched the scenes differently depending on the target category. When searching for a microwave (leftmost four panels) the model and behavioral searchers tended to look at counter-tops, but when searching for a clock (rightmost four panels) they tended to look higher up on the walls. Future work will more fully explore the potential to learn and predict these effects of scene context on search.

We directly compared the model and participant search behavior in several analyses of the test data. This comparison occurred on an image-by-image and fixation-by-fixation basis, but was limited to the first six movements of gaze. We introduced this limitation on the number of search saccades to reduce model computation time, but believe that it is justified given the clear adequacy of the first six saccades in revealing the goal-directed behavior of interest, as shown in Figure~\ref{fig:cumprob}. 

\begin{figure}[!hbt]
\centering
\includegraphics[width=1\textwidth,height=0.95\textheight,keepaspectratio]{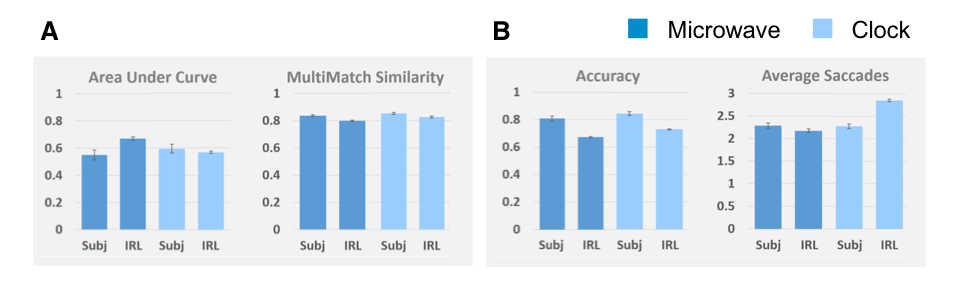}
\caption[short]{Model Performance Results. (A) Left: Success in predicting FDMs by the IRL model and a Subject model using the Area-Under-the-Curve metric. Right: Corresponding predictions of search scanpaths using average MultiMatch similarity. (B) Left: Proportion of trials in which the target was fixated in the first six saccades (fixated-in-6 accuracy). Right: Average number of saccades to the target on the fixated-in-6 trials. Note that in the (A) plots “Subj” refers to a Subject model whereas in the (B) plots "Subj" refers to behavioral data.}
\label{fig:modelperformance}
\end{figure}

Figure~\ref{fig:modelperformance}A (left plot) shows that the model was able to predict the behavioral FDMs for microwave and clock targets, using an AUC metric where the scale is between 0 and 1 and higher values indicate better predictive success. We also include Subject models, computed using the leave-one-out method, to obtain a practical noise limit on a model’s ability to predict group behavior~\cite{bylinskii2018different}. This analysis shows that the IRL model was able to predict the spatial distribution of behavioral fixations in the test images as well as could be expected based on variability among the participants in their search behavior. FDMs, however, are purely spatial, but search fixations are also made over time, ultimately producing a scanpath. Because the IRL model also makes sequences of fixations, we were able to compare its 6-saccade scanpaths to the 6-saccade scanpaths from the behavioral searchers (right plot). Based on average MultiMatch similarity~\cite{dewhurst2012depends}, excluding the fixation duration component, the model did a very good job in predicting the spatio-temporal sequences of fixations made by the behavioral searchers in their first six saccades, again as well as could be expected from the behavioral data, and it did this for both microwave and clock targets. 

We also analyzed search accuracy and the number of saccades that were made during search. Note that “Accuracy” refers here to the proportion of trials in which the target was fixated in the first six eye movements (fixated-in-6 accuracy), and “Avg Saccades” refers to the mean number of saccades needed to find the target on the accurate fixated-in-6 trials. Figure~\ref{fig:modelperformance}B shows that the model was slightly less successful than behavioral searchers in locating targets within six saccades (left plot), but when it did find the target it tended to do so about as efficiently as our participants, needing only about half a fixation more in the case of clocks (right plot). Chance fixated-in-6 accuracy is less than .25, based on a shuffling of eye data and images within each participant, and this is far lower than fixated-in-6 accuracy for the IRL model (microwave: t(58) = -31.74, p = 2.34e-38 <.001, Cohen's d = 8.20; clock: t(58) = -75.87, p = 9.73e-60 <.001, Cohen's d = 19.59). But perhaps the clearest measure of search efficiency is the cumulative probability of target fixation over saccades. As indicated by the dashed lines in Figure~\ref{fig:cumprob}, the model's search efficiency, although generally lower (reflecting the difference in fixated-in-6 accuracy), was strongly guided to targets much like the participants' search behavior. We interpret this as meaning that the IRL model learned goal-specific attention control, as measured by a gold-standard metric.

\section*{Discussion}
Models of search behavior have traditionally aimed at describing relatively coarse patterns (e.g., set-size effects) in highly simplified contexts~\cite{wolfe1994guided, zelinsky2008theory}, limitations that were imposed by a reliance on handcrafted features to create a guidance signal. In this study we adopted the radically different approach of training a model simply on many observations of search behavior, and showed that the Policy learned by this model predicted multiple overt measures of  goal-directed attention control. The success of these predictions is significant in that it requires a re-setting of the goal posts with respect to model evaluation. While once computational methods limited attention models to fitting patterns of search data in simple contexts, with deep networks it is possible to predict individual fixations made in the search for categories of objects in realistic scenes. 

Training this model required creating the Microwave-Clock Search dataset, which is among the only datasets of goal-directed attention (search fixations) large enough to train deep-network models. We encourage people to download this dataset from~\url{https://you.stonybrook.edu/zelinsky/datasetscode/} and use it in their own predictive-modeling work, citing this publication. Our hope is that the availability of this dataset will promote greater model development and comparison, which given the pace of recent advances might meaningfully advance the understanding of goal-directed attention control. 

The visual search for an object category is a goal-directed behavior of unique importance, shared by pigeons and people and most species in between. Because of its fundamental role in survival, search is likely to use the most basic of control processes—reward.~\cite{anderson2013value} Using the MCS dataset and Inverse-Reinforcement Learning, we showed that the target-specific reward functions learned by our model predicted the goal-directed fixations made by new people searching new images for the learned target categories. Machine learning has made it possible to learn the reward functions underlying goal-directed attention control. In ongoing work we are expanding our search dataset to 18 target categories so as to begin characterizing how reward functions vary over common real-world objects and to more fully explore scene context effects. In future work we also plan to manipulate different types of reward used in training, and apply IRL to questions in individual-difference learning.

\section*{Data availability}
The dataset described in this paper is available in the \textit{Microwave-Clock Search (MCS) dataset} repository: \url{https://you.stonybrook.edu/zelinsky/datasetscode/}.

\bibliography{IRL2020}

\section*{Acknowledgements}
We would like to thank the National Science Foundation for their generous support through award IIS-1763981, and members of the EyeCog Lab for their help with data collection and invaluable feedback. 

\section*{Author contributions}
M.H., D.S., and G.J.Z. conceptualized the research; H.A., Y.C., and G.J.Z. collected the dataset, L.H., Z.Y., D.S., and M.H. implemented the model.  All authors analyzed and interpreted data, but especially Z.Y., Y.C., L.H., and S.A.  G.J.Z., Y.C., S.A., and H.A. wrote the paper. 

\section*{Additional information}

\textbf{Competing interests}. 

The authors declare no competing interests.

\end{document}